**A Transferable Pretrained 3D Model for Impedance Inversion with Sparse-Well Prior Conditioning**

Gen Li [a], Zhaoyun Zong [a*], Diqiong Jiang [b], Man Tang [a]

[a] *State Key Laboratory of Deep Oil and Gas, China University of Petroleum (East China), Qingdao 266580, China.*

[b] *College of Computer Science and Technology, China University of Petroleum (East China), Qingdao 266580, P. R. China.*

**A Transferable Pretrained 3D Model for Impedance Inversion with Sparse-Well Prior Conditioning**

## Abstract

Deep-learning-based seismic impedance inversion commonly requires retraining for each target survey, which may generalize poorly beyond the conditioning locations when well control is extremely sparse. We propose a transferable three-dimensional inversion framework that separates offline learning from target-survey deployment. A masked autoencoder first learns seismic structural representations from unlabeled multi-survey data, after which the frozen encoder is incorporated into an inversion network trained on synthetic geological models with sparse-well-derived low-frequency priors. During deployment, the complete inversion network remains fixed; the available conditioning well is used only to construct the low-frequency prior and align the global impedance range. Experiments on the Overthrust model show that the proposed method achieves the lowest full-volume NRMSE and reduces the P95 trace-wise NRMSE from 0.1310 to 0.1074 relative to the best target-trained baseline, indicating fewer localized large errors under different well locations. In a field survey with one conditioning well and three blind wells, it reduces the mean blind-well NRMSE by 20.95% relative to the target-trained U-Net while maintaining the highest correlation. These results demonstrate that the proposed framework transfers useful impedance information beyond the conditioning location with low deployment cost and without retraining the inversion network.

**Keywords:** Seismic impedance inversion; Cross-survey transfer; Self-supervised pretraining; Sparse-well conditioning; Low-frequency prior.

## 1 Introduction

Seismic impedance inversion is an important tool for reservoir characterization and lithological interpretation because it transforms seismic observations into a subsurface property closely related to rock properties (Treitel and Lines, 2001; Zong et al., 2017, 2024; Chen et al., 2024a, 2024b). Deep-learning methods provide powerful nonlinear modeling capabilities for this task, but most existing models are developed within individual surveys. The relationships they learn can depend strongly on the geological background represented by the target-survey training data. Consequently, applying a model to a new survey commonly requires the inversion network to be retrained or fine-tuned using local well log (Meng et al., 2024). When such well information is extremely limited, however, supervised target-survey training can be dominated by the available well locations and learn a location-specific seismic–impedance relationship that may not generalize reliably to laterally different regions (Di et al., 2025). This study therefore investigates whether seismic representations and inversion knowledge learned offline can be transferred to a new survey under extremely sparse well control without target-specific optimization of the inversion network.

Existing supervised inversion approaches include convolutional neural networks (Das et al., 2019), recurrent neural networks (Alfarraj and AlRegib, 2018), hybrid models (Song et al., 2022), U-Net variants (Ning et al., 2023; Chakraborty et al., 2024), and Transformer-based models (Li et al., 2024). Multidimensional and multi-trace methods further incorporate neighboring seismic information to improve lateral continuity (Guo et al., 2021; Wu et al., 2021; Zhang et al., 2026), while joint learning across multiple datasets has also been explored to improve generalizability (Mustafa et al., 2021). Despite their architectural differences, these methods generally learn the seismic-impedance mapping from target-survey well logs. Their predictions therefore remain sensitive to the number, spatial distribution, and geological representativeness of the available wells, particularly when well control is extremely sparse (Zhang et al., 2022).

Semi-supervised and transfer-learning strategies reduce the demand for labeled well logs by incorporating unlabeled target seismic data, generated training samples, or synthetic pretraining (Alfarraj and AlRegib, 2019; Meng et al., 2024; Li et al., 2025; Su et al., 2026). However, these approaches generally retain a target-survey optimization stage in which the inversion network is trained or adapted using survey-specific information.

Self-supervised pretraining offers a possible way to move part of the learning process from individual target surveys to a reusable offline stage. By reconstructing masked seismic data or solving other label-free objectives, a model can learn reflection continuity, structural morphology, and three-dimensional spatial relationships from unlabeled seismic volumes (Dou and Li, 2024; Han et al., 2024; Sheng et al., 2025). Existing MAE-based three-dimensional inversion has demonstrated the benefit of pretrained representations under limited labels, although its downstream inversion mapping is still learned using target-survey well logs (Dou and Li, 2024). Beyond self-supervised pretraining, diffusion-based inversion has incorporated synthetic impedance knowledge and three-dimensional constraints through iterative posterior sampling (Liu et al., 2026). Existing approaches therefore still involve either target-specific downstream learning or iterative inference. This motivates learning both the seismic representation and the conditional inversion mapping offline and transferring them to a target survey as a frozen feed-forward model under extremely sparse well control.

A low-frequency impedance model provides a natural source of target-survey information (Zong et al., 2018; Sun and Zong, 2024). It supplies the slowly varying impedance background that cannot be reliably recovered from band-limited seismic data and allows the seismic observations to constrain relative variations and interfaces (Latimer et al., 2000; White and Zabihi Naeini, 2017). Alternative learning-based methods have sought to recover missing bandwidth through seismic low-frequency extrapolation, reflectivity decomposition, or seismic-envelope information (Torres and Sacchi, 2023; Cong et al., 2024; Dixit et al., 2025). In field inversion, however, low-frequency impedance information is commonly constructed from impedance logs, interpreted horizons, and spatial interpolation (Hansen et al., 2008). Under sparse well control, the lateral representativeness of this background depends on well placement and geological complexity. Treating an ideal low-frequency model as known throughout the target area can therefore lead to an overly optimistic assessment of inversion performance. A transferable inversion framework should therefore explicitly account for sparse-well-derived priors with realistic lateral uncertainty during both training and evaluation.

These considerations motivate separating computationally expensive representation and inversion learning from target-survey deployment. Transferable knowledge can be learned offline from multi-survey seismic data and controllable synthetic models, while target seismic data and a sparse-well-derived low-frequency prior provide survey-specific information at deployment. This design avoids fitting a new nonlinear seismic-impedance mapping from extremely limited target labels.

Following this strategy, we develop a transferable three-dimensional impedance inversion framework comprising two offline learning stages. In Stage I, a three-dimensional masked autoencoder is trained on seismic volumes from multiple survey areas to learn reusable structural representations. In Stage II, the pretrained encoder is incorporated into a conditional inversion network trained on synthetic geological models with sparse-well-derived low-frequency priors. The network predicts an impedance correction relative to the supplied prior, while dynamic gated fusion controls the injection of prior features according to the seismic and low-frequency information. An auxiliary branch estimates an equivalent Ricker dominant frequency and introduces convolutional consistency during offline training.

For a target survey, all offline-trained components remain fixed. The available conditioning well is used only to construct the low-frequency prior and determine two global affine-normalization parameters. This deterministic adjustment aligns the global impedance scale and offset without updating the network or learning a new target-specific seismic-impedance mapping.

The main contributions of this study are summarized as follows:

1. A transferable three-dimensional inversion framework is developed by combining multi-survey masked seismic modeling with offline synthetic impedance inversion training. The resulting seismic representations and inversion knowledge can be reused in a target survey without retraining the complete inversion network.

2. Sparse-well-derived priors with lateral uncertainty are explicitly incorporated into offline inversion training. Residual prediction and dynamic gated fusion enable the decoder to combine three-dimensional seismic information with low-frequency priors whose lateral representativeness varies away from the conditioning wells.

3. Controlled comparisons in which all methods receive the same one-well information show that the transferred prediction preserves lateral impedance information beyond the conditioning location and reduces localized large errors relative to target-trained baselines, without target-specific network optimization.

The evaluation includes an independent Stage-I reconstruction case, five fixed one-well configurations of the Overthrust model, and a field survey containing one conditioning well and three blind wells. All six target-trained residual-learning baselines receive the same seismic data, conditioning well, and low-frequency prior as the proposed method.

## 2 Methodology

### 2.1 Problem formulation and framework overview

Post-stack seismic impedance inversion aims to recover the subsurface impedance model from seismic data. Let the input seismic volume be denoted as $S$, and the target impedance volume as $I$. Under the approximation of the convolutional model, a seismic record can be represented as the band-limited response obtained by convolving the reflection coefficient series with a seismic wavelet (Robinson, 1967):

$$s(t) = w(t) * r(t) + n(t) \tag{1}$$

where $w(t)$ is the seismic wavelet, $r(t)$ is the reflection coefficient series, and $n(t)$ denotes the noise term. For the impedance model $I$, the reflection coefficient between two adjacent samples is given by (Russell, 1988):

$$r(t) = \frac{I(t+\Delta t) - I(t)}{I(t+\Delta t) + I(t)} \tag{2}$$

Equations (1) and (2) indicate that band-limited seismic data primarily constrain impedance contrasts, whereas the slowly varying absolute impedance background remains weakly constrained. The inversion is therefore conditioned on a target-survey low-frequency impedance prior $I_L$, and the network predicts a correction relative to this background:

$$\hat{I} = I_L + \Delta\hat{I}_\theta(S, I_L) \tag{3}$$

where $\Delta\hat{I}_\theta$ is estimated from both seismic and low-frequency features. The residual formulation in Eq. (3) benefits from a reliable low-frequency background. In practice, however, $I_L$ may contain spatial uncertainty when constructed from sparse wells. The treatment of this uncertainty is described in Sections 2.3 and 2.4.

The framework comprises two offline learning stages followed by frozen target-survey deployment (Fig. 1). In Stage I, a three-dimensional masked autoencoder is trained using unlabeled seismic volumes from multiple surveys. Its encoder learns spatial representations of reflection continuity and structural morphology without using impedance labels. In Stage II, the pretrained encoder remains frozen and is incorporated into an inversion network trained offline on synthetic geological models. The Stage-II network combines the transferred seismic representations with low-frequency priors constructed from different numbers and layouts of sparse wells. An auxiliary branch additionally estimates an equivalent Ricker dominant frequency and introduces convolutional forward consistency during offline training.

For a new target survey, the target seismic volume and sparse-well low-frequency prior are processed by the offline-trained three-dimensional network, with all network parameters kept fixed. Once the frozen prediction for the entire survey has been obtained, it is aligned with the target impedance range using the conditioning-well affine normalization described in Section 2.7.3. This deterministic adjustment requires no target-specific network optimization.

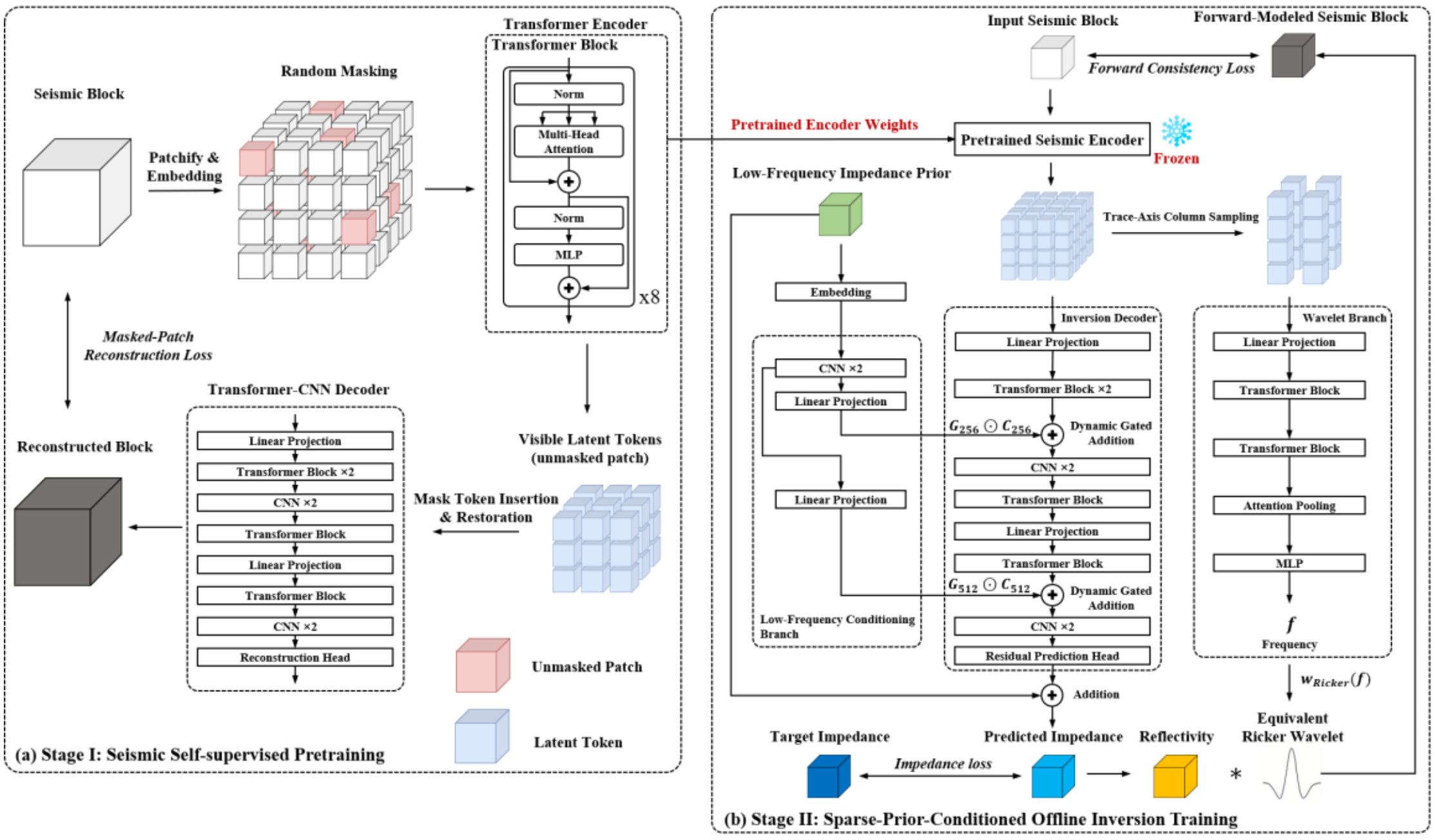

**Fig.1.** Overview of the proposed two-stage offline learning framework. (a) Stage I:

masked seismic representation learning using unlabeled multi-survey seismic data; (b) Stage II: sparse-prior-conditioned offline inversion training with a frozen seismic encoder and dominant-frequency-based convolutional consistency.

## 2.2 Stage-I masked seismic representation learning

### 2.2.1 Three-dimensional patch masking

Stage I follows the masked-autoencoding principle (He et al., 2022) but operates directly on three-dimensional seismic blocks. An input block of size 120×120×120 is divided into non-overlapping 8×8×8 patches. A three-dimensional convolution whose kernel and stride equal the patch size maps the patches to a 15×15×15 token grid, giving 3,375 patch tokens. Absolute positional embeddings are added to retain the spatial arrangement of the tokens.

For each input block, 80% of the patch tokens are randomly masked, and only the remaining 20% are processed by the encoder. The encoder therefore has no direct access to the amplitudes within the masked patches and must infer their content from the visible three-dimensional reflection context. Processing only the visible tokens also reduces the self-attention computation during pretraining.

### 2.2.2 Encoder and reconstruction decoder

The encoder contains eight Transformer blocks with an embedding dimension of 512, eight attention heads, and an MLP expansion ratio of four (Vaswani et al., 2017). A learnable classification token is included to aggregate block-level context. The visible tokens and the classification token are passed through the Transformer blocks and a final layer-normalization operation.

The reconstruction decoder is used only in Stage I. The encoded visible tokens are first projected from 512 to 256 dimensions. Learnable mask tokens are inserted at the missing positions, the original token order is restored, and decoder positional embeddings are added. Two 256-dimensional Transformer blocks first establish long-range interactions between the visible and masked positions. Two 3×3×3 convolutional refinement blocks then operate on the restored token grid to improve local spatial continuity. After one additional 256-dimensional Transformer block, the features are expanded to 512 dimensions and processed by one Transformer block and two further 3×3×3 convolutional refinement blocks. A linear reconstruction head maps each decoded token back to its original seismic patch.

The reconstruction loss is evaluated only at the masked positions:

$$L_{\text{rec}} = \frac{1}{|M|P} \sum_{i \in M} \left\| \mathbf{x}_i - \mathbf{x}_i \right\|_1, \qquad P = 8^3 \tag{4}$$

where $M$ is the set of masked patch indices, $P$ is the number of samples in each patch, and $x_i$ and $\hat{x}_i$ are the original and reconstructed amplitude vectors of the $i$-th patch, respectively. This masked raw-amplitude term is the sole Stage-I training objective. After Stage I, the reconstruction decoder is discarded, and only the pretrained encoder is transferred to the subsequent inversion stage. Given an unmasked 120×120×120 seismic block, the encoder produces a CLS token and 3,375 patch tokens, each with a dimension of 512. The patch tokens are further reshaped into a 512×15×15×15 latent feature grid for Stage II.

## 2.3 Stage-II training data and sparse low-frequency priors

### 2.3.1 Synthetic seismic generation

Stage II learns the conditional mapping in Eq. (3) using synthetic geological models for which impedance and reflection coefficients are available. During training, A 120×120×120 block is randomly cropped from each selected parent model. The seismic input is synthesized from the reflection-coefficient volume using a stationary zero-phase Ricker wavelet:

$$w(t; f) = \left[1 - 2(\pi f t)^2\right] e^{-(\pi f t)^2} \tag{5}$$

where $f$ denotes the dominant frequency. The discretized wavelet is normalized to unit $L2$ energy before convolution. During training, $f$ is sampled from 25 to 50 Hz, with sampling probabilities of 0.4, 0.4, and 0.2 assigned to the intervals 25-32, 32-40, and 40-50 Hz, respectively. No phase rotation or time shift is introduced. Additive white Gaussian noise is independently added to each synthetic seismic block, with the volume-level signal-to-noise ratio sampled uniformly between 0 and 20 dB.

Each noisy seismic block is normalized by its 99th percentile of absolute amplitude. For impedance normalization, the 1st and 99th percentiles of the complete parent impedance model are used as the lower and upper bounds. Both the target impedance and its low-frequency prior are transformed using the same bounds, such that the two percentile values correspond to -1 and 1, respectively. During training, the bounds are independently perturbed by up to 2% of their percentile span to reduce sensitivity to a fixed impedance range. The corresponding bounds are retained to convert the predicted impedance back to physical units.

2.3.2 Sparse-well prior construction

A low-frequency prior obtained by filtering a complete impedance volume would provide unrealistically accurate lateral background information for a sparse-well deployment scenario. Therefore, most Stage-II training priors are constructed exclusively from sparse vertical impedance logs.

For each synthetic parent impedance volume, $N_w$ is selected from $\{1,3,5,7,9\}$, with the five cases approximately balanced among the sparse-prior samples within each epoch. Well coordinates are generated over the complete lateral extent of the parent model before crop extraction, using a distance-constrained spatially dispersed sampling procedure. The admissible region excludes a 5% margin along each lateral boundary, and a target minimum pairwise spacing of 10% of the diagonal is used. Candidate coordinates are sampled randomly, after which spatially separated candidates are preferentially selected. Training layouts vary between epochs but are deterministically generated from the global seed, epoch, sample index, and well count. The validation layouts remain fixed so that changes in validation loss reflect model performance rather than different well realizations.

Each selected impedance log is low-pass filtered at 10 Hz using a fourth-order zero-phase Butterworth filter. Let $l_j(t)$ denote the filtered log at location $(x_j, y_j)$. The low-frequency prior within a sampled crop is constructed by two-dimensional inverse-distance weighting (IDW) at each temporal sample:

$$I_L^{IDW}(x,y,t)=\frac{\sum_{j=1}^{N_w} d_j(x,y)^{-p} I_j(t)}{\sum_{j=1}^{N_w} d_j(x,y)^{-p}},$$
$$d_j(x,y)=\sqrt{(x-x_j)^2+(y-y_j)^2} \tag{6}$$

where $p = 2$. At an exact well location, the corresponding filtered log is retained directly. In the single-well case, IDW reduces to lateral replication of the filtered well log, providing no lateral structural information. With multiple wells, lateral variations arise from the spatially varying relative contributions of the available logs. The construction of $I_L^{IDW}$ uses no impedance samples at non-well locations; the complete impedance crop is used only as the Stage-II supervision target.

Sparse IDW priors and complete-volume filtered priors are used in a fixed 4:1 ratio. Thus, 80% of the training samples in each epoch use $I_L^{IDW}$, whereas the remaining 20% use the complete 10-Hz filtered impedance volume. This schedule emphasizes deployment-like sparse priors while retaining examples in which the supplied low-frequency background is reliable. Together with the dynamic gate introduced below, it enables input-dependent fusion of seismic and low-frequency condition features under different prior conditions.

## 2.4 Sparse-prior-conditioned inversion network

The Stage-II inversion pathway comprises the frozen seismic encoder, a low-frequency conditioning branch, and a newly initialized inversion decoder. The auxiliary dominant-frequency branch is described in Section 2.5.

### 2.4.1 Frozen seismic encoder and low-frequency conditioning branch

The encoder transferred from Stage I remains fixed throughout Stage II and receives a complete, unmasked seismic block. For an input seismic block $S$, the encoder output is

$$\left(z_{cls}, Z_S\right)=E_{\theta_0}(S) \tag{7}$$

where $E_{\theta_0}$ is the frozen encoder with parameters $\theta_0$, $z_{cls}$ is the CLS token, and $Z_S \in$

$\mathbb{R}^{3375\times512}$ contains the patch tokens corresponding to the $15\times15\times15$ spatial grid.

The normalized low-frequency prior $I_L$ is mapped onto the same grid using a three-dimensional convolution with a kernel and stride of $8\times8\times8$. Two convolution blocks refine the resulting 128-channel features. After flattening, two linear projections and layer normalization produce:

$$\{C_{256}, C_{512}\} = C_{\psi}(I_L) \tag{8}$$

where $C_{\psi}$ denotes the low-frequency conditioning branch, $C_{256} \in \mathbb{R}^{3375\times256}$, and $C_{512} \in \mathbb{R}^{3375\times512}$. The condition tokens are spatially aligned with the seismic patch tokens, allowing the low-frequency information to be introduced at corresponding token locations during decoding.

### 2.4.2 Inversion decoder and dynamic gated fusion

The inversion decoder is trained from random initialization and does not inherit parameters from the Stage-I reconstruction decoder. It uses Transformer blocks for long-range interactions and three-dimensional convolutional blocks for local refinement. The seismic features are processed at dimensions of 256 and 512, with the corresponding low-frequency condition tokens injected before the convolutional refinement at each level.

At decoder level $k \in \{256,512\}$, let $Z_k$ denote the seismic-decoder patch features and $C_k$ the aligned low-frequency condition features. Their dynamic gate is calculated as

$$G_k = \sigma(W_{g,k}[Z_k \,\|\, C_k] + b_{g,k}) \tag{9}$$

where $\|$ denotes feature concatenation and $\sigma$ is the sigmoid function. The fused features are then obtained as

$$\tilde{Z}_k = Z_k + G_k \odot C_k \tag{10}$$

where $\odot$ denotes element-wise multiplication. The gate $G_k$ has the same token and channel dimensions as $Z_k$ and $C_k$. It is calculated jointly from the current seismic-decoder and low-frequency condition features and therefore varies across input samples, token locations, and feature channels. This mechanism makes the contribution of the low-frequency condition features input dependent rather than controlling it with a fixed scalar or channel-wise coefficient.

The final tokens are mapped to $8 \times 8 \times 8$ impedance-correction patches and reassembled into a normalized volume:

$$\hat{I} = I_L^n + \Delta\hat{I}_\theta^n \tag{11}$$

where $n$ denotes normalized impedance. The low-frequency prior therefore supplies both the explicit output background and the condition features injected at the two decoder levels.

## 2.5 Dominant-frequency estimation and convolutional consistency

The auxiliary branch estimates the equivalent Ricker dominant frequency used in Stage-II seismic generation. From the $15\times15\times15$ latent grid produced by the frozen encoder, four token columns are selected: the central column and three randomly sampled noncentral columns. Each column contains 15 tokens, which are projected from 512 to 256 dimensions and combined with a learnable query token and the encoder CLS token. Two Transformer blocks encode each column, and attention pooling combines the four query descriptors. A two-layer GELU prediction head produces $u_f \in [0,1]$, which is converted to frequency by

$$f = f_{\min} + u_f (f_{\max} - f_{\min}) \tag{12}$$

Fixed column locations are used during validation and inference to ensure deterministic estimates. An equivalent Ricker wavelet $w(t;\hat{f})$ is constructed from the predicted frequency. The impedance prediction is converted back to physical units, and its reflection coefficients are calculated using Eq. (2). The corresponding seismic response is

$$\hat{s}(x,y,t) = w(t;\hat{f}) * r\left(\hat{I}(x,y,t)\right) \tag{13}$$

where $*$ denotes convolution along the temporal axis.

The modeled and input seismic traces are standardized independently before evaluating the convolutional-consistency loss. This differentiable path provides auxiliary consistency regularization during Stage II, while the predicted dominant frequency can also be reported as a model-derived diagnostic during target-survey application.

## 2.6 Joint Stage-II objective

The Stage-II objective combines impedance-domain losses with dominant-frequency supervision and convolutional consistency. In Eqs. (14)-(20), $\hat{I}$ and $I$ denote the normalized prediction and target, respectively, and all $L_1$ norms represent mean absolute errors.

The pointwise value loss is defined using the Smooth $L_1$ function with a transition width of $\beta = 1$:

$$L_{val} = \mathrm{SmoothL1}_{\beta=1}(\hat{I}, I) \tag{14}$$

To reduce sensitivity to the block-level mean and scale, a relative-shape loss compares the independently standardized prediction and target:

$$L_{rel} = \left\| \frac{\hat{I} - \mu_{\hat{I}}}{\sigma_{\hat{I}}} - \frac{I - \mu}{\sigma_I} \right\|_1 \tag{15}$$

where $\mu_{\hat{I}}$, $\sigma_{\hat{I}}$, $\mu_I$, and $\sigma_I$ are calculated separately for each three-dimensional block. This term emphasizes the relative spatial and temporal impedance patterns rather than their block-level amplitude ranges.

A vertical-gradient loss is used to preserve impedance contrasts at geological interfaces:

$$L_{grad} = \left\| \nabla_t \left( \frac{\hat{I} - \mu_{\hat{I}}}{\sigma_I} \right) - \nabla_t \left( \frac{I - \mu_I}{\sigma_I} \right) \right\|_1 \tag{16}$$

where $\nabla_t$ denotes the finite difference along the temporal axis. In contrast to Eq. (15), both gradients are scaled by the target standard deviation $\sigma_I$. The loss therefore retains differences in interface magnitude while reducing their dependence on the impedance range of individual training models.

Trace-level waveform similarity is measured using the Pearson correlation coefficient (PCC):

$$L_{PCC} = 1 - \frac{1}{N_{tr}} \sum_{q=1}^{N_{tr}} PCC(\hat{I}_q, I_q) \tag{17}$$

where $q$ indexes the impedance traces and $N_{tr}$ is the number of traces in the batch.

The standardized impedance volumes are additionally flattened into traces and decomposed into five complementary temporal bands using moving-average kernels of 11, 21, 41, and 61 samples. The weighted L1 differences between corresponding bands define the multiband loss $L_{mb}$, with band weights {0.05,0.10,0.20,0.35,0.30} from low to high temporal detail. An amplitude-weighted spectral phase-alignment loss, denoted by $L_{\phi}$, is evaluated over the selected intermediate bands.

For dominant-frequency supervision, the synthetic frequency label is normalized to [0,1]:

$$u_f = \frac{f - f_{\min}}{f_{\max} - f_{\min}}, \qquad \mathcal{L}_f = \mathrm{SmoothL1}(\hat{u}_f, u_f) \tag{18}$$

where $\hat{u}_f$ is the output of the auxiliary branch. For the modeled seismic response in Eq. (13), the convolutional-consistency loss is

$$L_{\mathrm{cf}} = 0.8L^{s}_{\mathrm{corr}} + L^{s}_{\mathrm{mb}} + 0.1L^{s}_{\phi} \tag{19}$$

where the superscript $s$ denotes losses evaluated between the trace-wise standardized modeled and input seismic data.

The complete Stage-II objective is:

$$\begin{aligned} L_{\mathrm{total}} = {} & 0.5L_{\mathrm{val}} + 0.5L_{\mathrm{rel}} + 0.2L_{\mathrm{grad}} + 0.2L_{\mathrm{corr}} + 0.3L_{\mathrm{mb}} \\ & + 0.01L_{\phi} + 0.10L_{f} + 0.02L_{\mathrm{cf}} \end{aligned} \tag{20}$$

Checkpoint selection is based on the minimum weighted composite validation loss across six fixed prior conditions: the complete 10-Hz filtered prior and the 1-, 3-, 5-, 7-, and 9-well IDW priors. Their weights are 0.20 and 0.16 for each IDW condition, respectively, matching the 4:1 training ratio and preventing checkpoint selection from being dominated by the complete-prior condition.

## 2.7 Target-survey deployment

### 2.7.1 Frozen target-survey prediction

A low-frequency impedance prior is first constructed for the target survey. In the synthetic experiments, the conditioning-well log is filtered and interpolated using Eq. (6). For the field survey, the filtered conditioning-well log is interpolated along interpreted seismic horizons. The target seismic volume and low-frequency prior are then normalized and divided into three-dimensional blocks.

All offline-trained components remain fixed during target-survey application. For each block, the normalized frozen prediction is

$$\hat{I}^{F,n} = F_{\theta^*}(S^n, I_L^n) = I_L^n + \Delta\hat{I}^n_{\theta^*}(S^n, I_L^n), \tag{21}$$

where $n$ denotes normalized quantities and $\theta^*$ contains the offline-trained parameters. No target-survey impedance label is used to update these parameters.

2.7.2 Full-survey prediction

The network operates on $120\times120\times120$ blocks. Larger volumes are processed using overlapping blocks, with the origins and overlap specified in the experimental setup. Predictions in overlapping regions are combined using separable edge-tapered weights and divided by the accumulated weight volume. This blending reduces block-boundary artifacts without modifying the network output. Once prediction over the entire survey is complete, the retained impedance normalization parameters are used to recover physical units.

2.7.3 Conditioning-well affine normalization

The frozen prediction may retain a systematic difference in impedance mean or scale after cross-survey transfer. This difference is corrected using statistics from the conditioning well. Let $I_w$ denote the measured impedance log and $\hat{I}_w^F$the full-survey frozen prediction extracted at the same location and time interval. The normalization parameters are

$$a = \frac{\sigma(I_w)}{\sigma(\hat{I}_w^F) + \epsilon}, \qquad b = \mu(I_w) - a\mu(\hat{I}_w^F) \tag{22}$$

and the final prediction is

$$\hat{I}^{N} = a\hat{I}^{F} + b \tag{23}$$

The same $a$ and $b$ are applied uniformly to the entire survey. This deterministic operation matches the prediction mean and standard deviation to the conditioning log without updating any network parameter or fitting a new seismic–impedance mapping. Blind wells are excluded from the calculation.

## 3 Experiments and results

The experimental evaluation comprises three parts. First, masked seismic reconstruction on held-out data is used to evaluate the representations learned in Stage I. Second, five single-well configurations of the Overthrust model are used to compare the frozen transferable prediction with target-trained residual-learning baselines supplied with the same low-frequency priors. Finally, a field survey containing one conditioning well and three blind wells is used to evaluate practical application under extremely sparse well control.

### 3.1 Experimental data and protocols

#### 3.1.1 Offline-training data and implementation

The Stage I seismic data originated from 34 survey areas worldwide (U.S. Geological Survey, 2014; dGB Earth Sciences, 2023). Spatial resampling produced 268 three-dimensional volumes: 244 from 31 areas formed the training set, while 24 from three held-out areas were excluded from training. The training manifest contained 124,769 blocks of size $120\times120\times120$. A fixed subset from one held-out area was used for validation and checkpoint selection. The Gulf of Mexico B-40H-89-LA survey analyzed in Section 3.2 was reserved as an independent reconstruction test case, and the field survey examined in Section 3.4 was also absent from Stage I training and validation. No impedance labels were used during Stage I.

The Stage II geological model data were generated using a customized implementation based on the open-source cigChannel framework (Wang et al., 2025). The framework was extended to generate three-dimensional stratigraphic models with randomized tilting, folding, localized fill, mound-like deposition, faulting, and P-wave velocity distributions. Under the constant-density assumption adopted in cigChannel, impedance was calculated as

$$I_p(\mathbf{x}) = \rho V_p(\mathbf{x}), \qquad \rho = 2.4\ \mathrm{g/cm^3} \tag{24}$$

The pooled 1st-99th percentile interval of the resulting impedance models was $5.21 \times 10^3 - 11.97 \times 10^3\ \mathrm{m/s \cdot g/cm^3}$. In total, 200 independently generated models of size 250×250×250 were obtained; 180 were used for training and 20 were reserved for validation. Random 120×120×120 blocks were cropped from these models during Stage II training.

Seismic synthesis and sparse-well prior construction followed Sections 2.3.1 and 2.3.2, respectively. Neither the Overthrust model nor the field-survey impedance labels were used for Stage II optimization or checkpoint selection. Representative data from the two offline-training stages are shown in Fig. 2.

Both stages were implemented in PyTorch and trained on a single NVIDIA V100S GPU using mixed-precision computation. AdamW was used with an initial learning rate of $1.0 \times 10^{-4}$ and a weight decay of $1.0 \times 10^{-2}$. Stage I was trained for at most 78000 optimization steps, with a batch size of 32 and a 2000-step warm-up followed by cosine decay. Stage II was trained for at most 13800 steps with a batch size of 8 and a 230-step warm-up. In both stages, the warm-up was followed by cosine learning-rate decay. Checkpoint selection in Stage II was based on the composite validation loss described in Section 2.6.

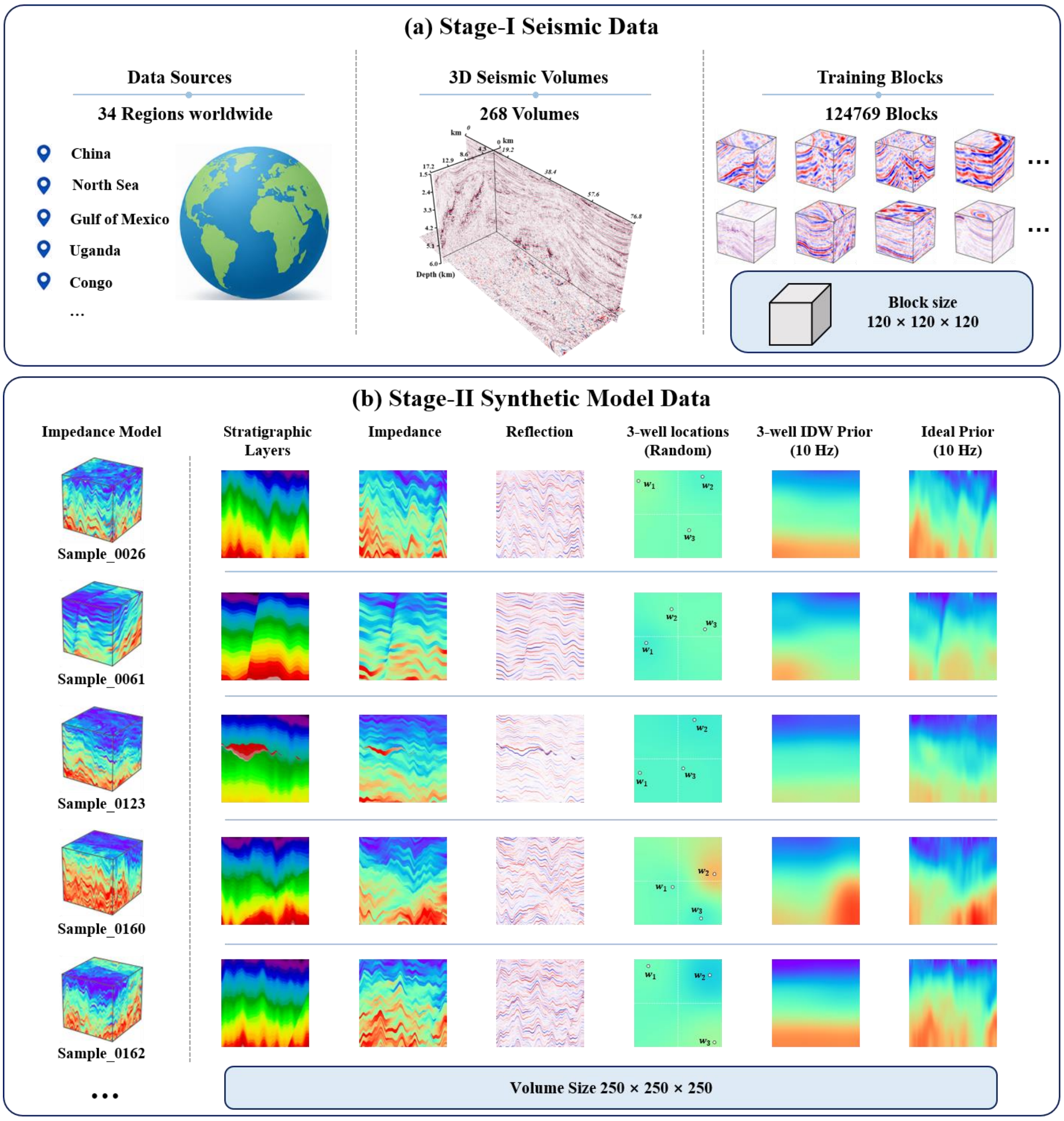

Fig. 2. Overview of the data used in the two offline-training stages. (a) Stage-I seismic data collected from 34 survey areas and divided into 120×120×120 blocks for masked seismic modeling; (b) Representative Stage-II synthetic geological model data. Each row corresponds to an independently generated 250×250×250 model.

### 3.1.2 Target datasets and well layouts

The synthetic test volume was a 360×360×360 subset of the Overthrust model (Aminzadeh et al., 1997), with a temporal sampling interval of 1 ms. Its reflection coefficients were convolved with a 30-Hz Ricker wavelet to generate the seismic input. This volume was excluded from offline training and checkpoint selection.

Five fixed single-well configurations were evaluated, with the conditioning wells located at lateral grid indices (60,60), (300,300), (300,60), (60,300), and (180,180). For each configuration, the conditioning log was low-pass filtered at 10 Hz and used to construct the low-frequency prior according to Eq. (6). With only one well, inverse-distance weighting reduces to lateral replication of the filtered log. The resulting prior therefore provides a vertical impedance trend but no lateral structural variation. No impedance samples outside the designated conditioning well were used for prior construction, target-specific baseline training, or affine normalization. Full-volume metrics were calculated over all 129600 lateral traces, and the results were summarized across the five configurations.

The frozen three-dimensional network was applied using an overlap of 40 samples and a stride of 80 samples along each axis. Overlapping block predictions were blended using the procedure described in Section 2.7.2.

### 3.1.3 Comparison methods and ablation settings

Six target-trained baselines were considered: a one-dimensional CNN, a GRU, a CNN-GRU hybrid, a one-dimensional U-Net, an attention U-Net, and a lightweight Transformer. Each received the same seismic trace and one-well low-frequency prior and predicted a correction to the prior. The prior was also reported as a no-training reference.

For each well configuration, the baselines were trained from random initialization using only the designated conditioning trace. Training used a learning rate of $1.0 \times 10^{-3}$, a weight decay of $1.0 \times 10^{-4}$, and at most 2,000 epochs. No additional validation well was used because, although excluded from gradient updates, its impedance labels would influence checkpoint or hyperparameter selection and thereby provide additional target-survey information. This would be inconsistent with the intended one-well setting, and the validation well could no longer be regarded as blind. All training settings were therefore fixed in advance, and the checkpoint with the lowest conditioning-well loss was retained.

The proposed network remained fully frozen. After full-volume prediction, the same conditioning-well affine normalization was applied to all learned methods using only the designated conditioning log.

Three ablations removed Stage-I pretraining, gated low-frequency feature injection, and dominant-frequency consistency, respectively. The remaining data and evaluation settings were unchanged. The baseline with the lowest mean NRMSE across the five Overthrust configurations was subsequently used in the field comparison.

3.1.4 Evaluation metrics

The impedance results were evaluated in physical units using normalized root mean square error (NRMSE), and the Pearson correlation coefficient (PCC):

$$\mathrm{NRMSE} = \frac{\sqrt{\frac{1}{N}\sum_{i=1}^{N}(\hat{I}_i - I_i)^2}}{I_{\max} - I_{\min}} \tag{25}$$

$$\mathrm{PCC} = \frac{\sum_{i=1}^{N}(\hat{I}_i - \bar{\hat{I}})(I_i - \bar{I})}{\sqrt{\sum_{i=1}^{N}(\hat{I}_i - \bar{\hat{I}})^2 \sum_{i=1}^{N}(I_i - \bar{I})^2}} \tag{26}$$

For each experiment, the target impedance range $I_{max} - I_{min}$ was calculated once from the evaluation target and shared by all methods. Spatial error concentration in the Overthrust model was further characterized using trace-wise NRMSE maps and their 95th percentile (P95). Stage-I reconstruction was evaluated using NRMSE, PCC and three-dimensional structural similarity (SSIM):

$$SSIM_{3D}(\hat{S}, S) = \frac{1}{M}\sum_{m=1}^{M}\frac{(2\mu_{\hat{S},m}\mu_{S,m} + C_1)(2\sigma_{\hat{S}S,m} + C_2)}{(\mu_{\hat{S},m}^2 + \mu_{S,m}^2 + C_1)(\sigma_{\hat{S},m}^2 + \sigma_{S,m}^2 + C_2)} \tag{27}$$

where $M$ is the number of local three-dimensional windows; $\mu$, $\sigma^2$, and $\sigma_{\hat{S}S}$ denote the corresponding local means, variances, and covariance; and $C_1$ and $C_2$ are stability constants.

## 3.2 Stage-I seismic reconstruction

The held-out B-40H-89-LA volume from the Gulf of Mexico was used to evaluate Stage-I reconstruction on an unseen survey. With a masking ratio of 80%, the encoder received only 20% of the patch tokens. As shown in Fig. 3, the reconstructed volume preserves the main reflection events and their three-dimensional continuity, with residual errors concentrated near discontinuities. The reconstruction achieved a NRMSE of 0.0089, a PCC of 0.9310 and an SSIM of 0.8969, suggesting that the Stage-I encoder effectively represents the principal spatial structures of seismic data not included in training.

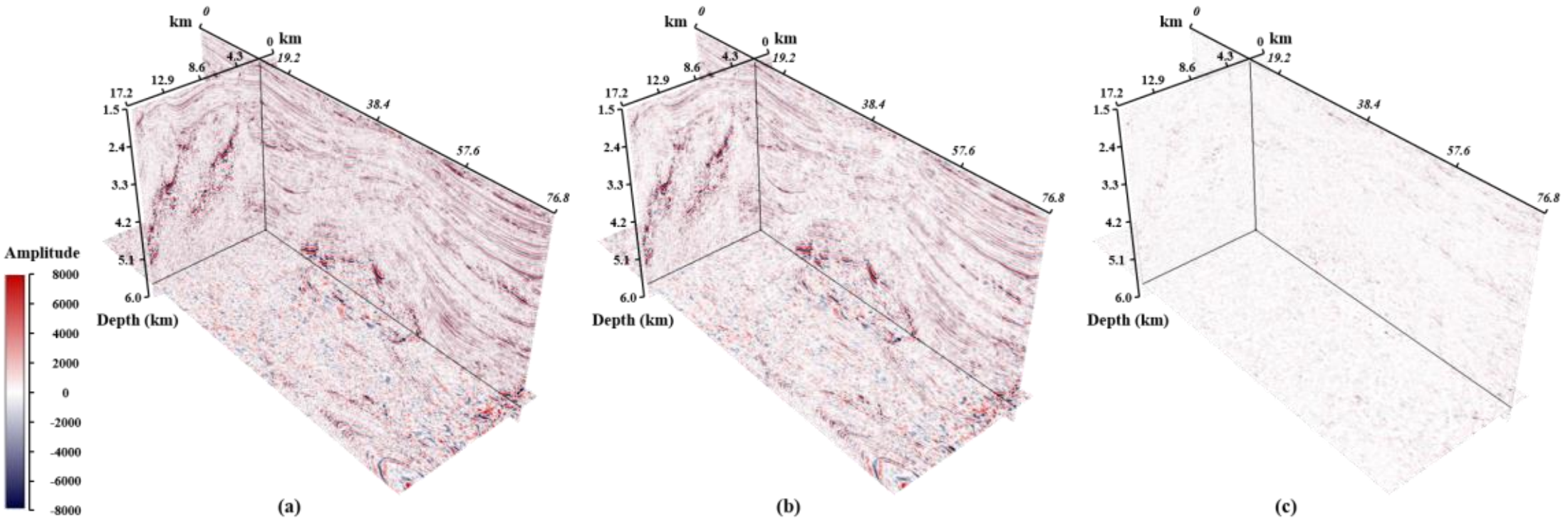

Fig. 3. Reconstruction results of seismic data. (a) Ground truth; (b) Reconstruction seismic; (c) Residual.

### 3.3 Overthrust experiments

Figure 4 presents the Overthrust target, synthetic seismic data, and five single-well low-frequency priors. Each of $W_1$ - $W_5$ defines an independent conditioning configuration. With only one well, inverse-distance weighting reduces to lateral replication of the filtered log. The priors in Figs. 4(d)-(h) therefore provide different vertical impedance trends but contain no lateral structural variation. The complete-volume filtered result in Fig. 4(b) is included only as an ideal low-frequency reference and was not used in the comparisons.

The proposed method was applied independently to the five configurations without updating its offline-trained parameters. After full-volume prediction, the same conditioning-well affine normalization was applied to the proposed method and all target-trained baselines. Table 1 reports the full-volume results averaged across the five configurations.

Among the target-trained baselines, U-Net achieved the lowest mean NRMSE of 0.0831 ± 0.0098. The proposed method obtained a lower NRMSE of 0.0814 ± 0.0057, together with the highest mean PCC of 0.9602. More notably, its P95 trace-wise NRMSE was 0.1074 ± 0.0093, whereas the mean P95 values of the six baselines ranged from 0.1310 to 0.1393. The smaller NRMSE variation and substantially lower P95 indicate that the proposed method is less sensitive to the conditioning-well location and produces fewer localized large errors without target-specific neural-network optimization. U-Net was subsequently selected for the field comparison as the best-performing target-trained baseline.

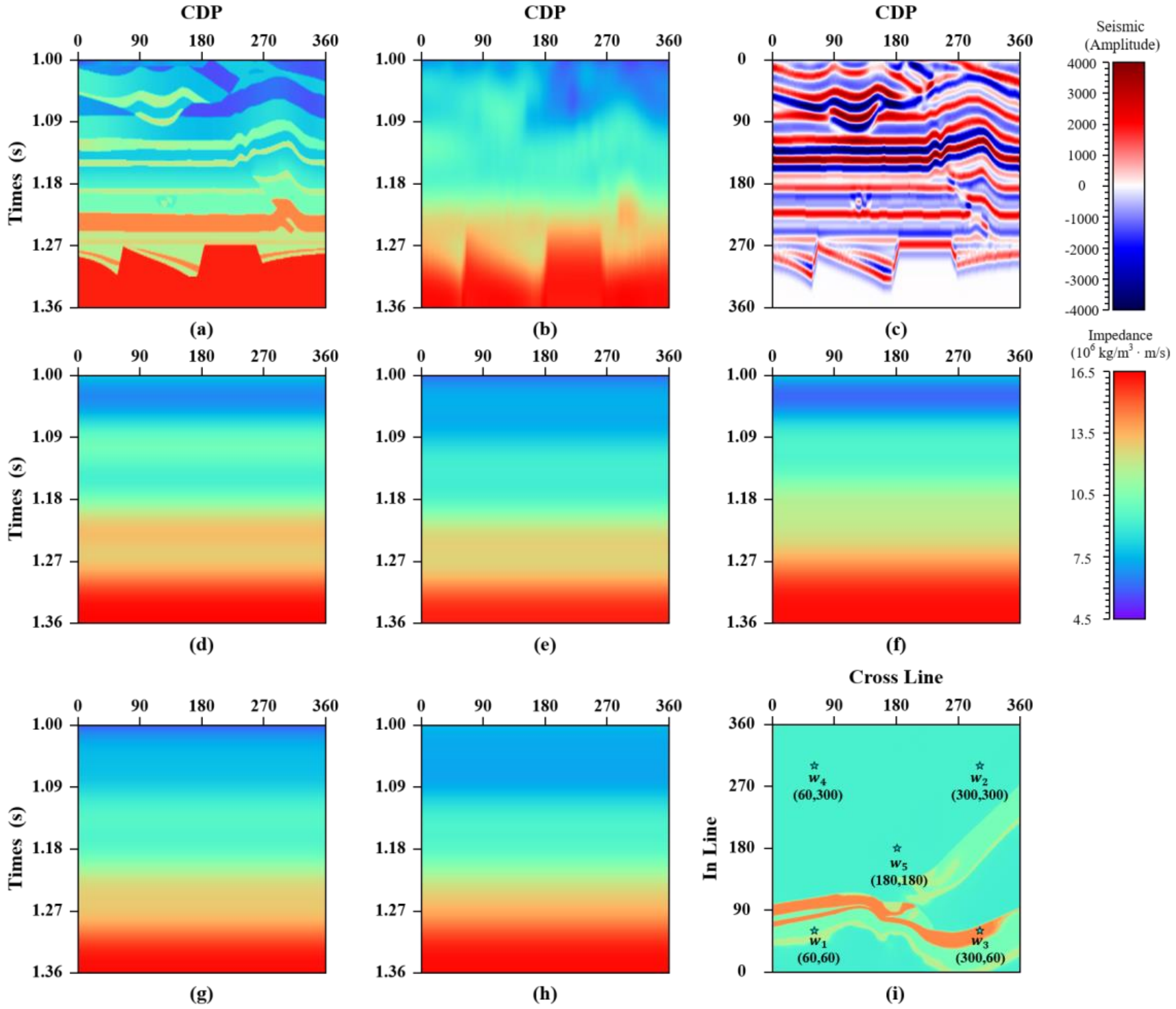

Fig. 4. Overthrust model and five single-well prior configurations. (a) Target impedance; (b) 10-Hz ideal low-frequency reference; (c) seismic data generated using a 30-Hz Ricker wavelet; (d)-(h) single-well 10-Hz IDW priors constructed from $W_1$ - $W_5$, respectively; (i) conditioning-well locations.

The trace-wise NRMSE maps in Fig. 5 show that the differences are not confined to the conditioning locations. The target-trained baselines produce localized high-error regions whose positions and extent vary with the selected well. In comparison, the proposed method maintains a more consistent lateral error distribution and fewer high-error regions across the five configurations, in agreement with its lower P95 value.

Across the 64 overlapping inference blocks, the auxiliary branch estimated an equivalent dominant frequency of 30.55 ± 1.09 Hz for the seismic volume generated using a 30-Hz Ricker wavelet. The block-averaged estimate differed from the prescribed frequency by 0.55 Hz, supporting its use as an equivalent dominant-frequency diagnostic under this synthetic condition.

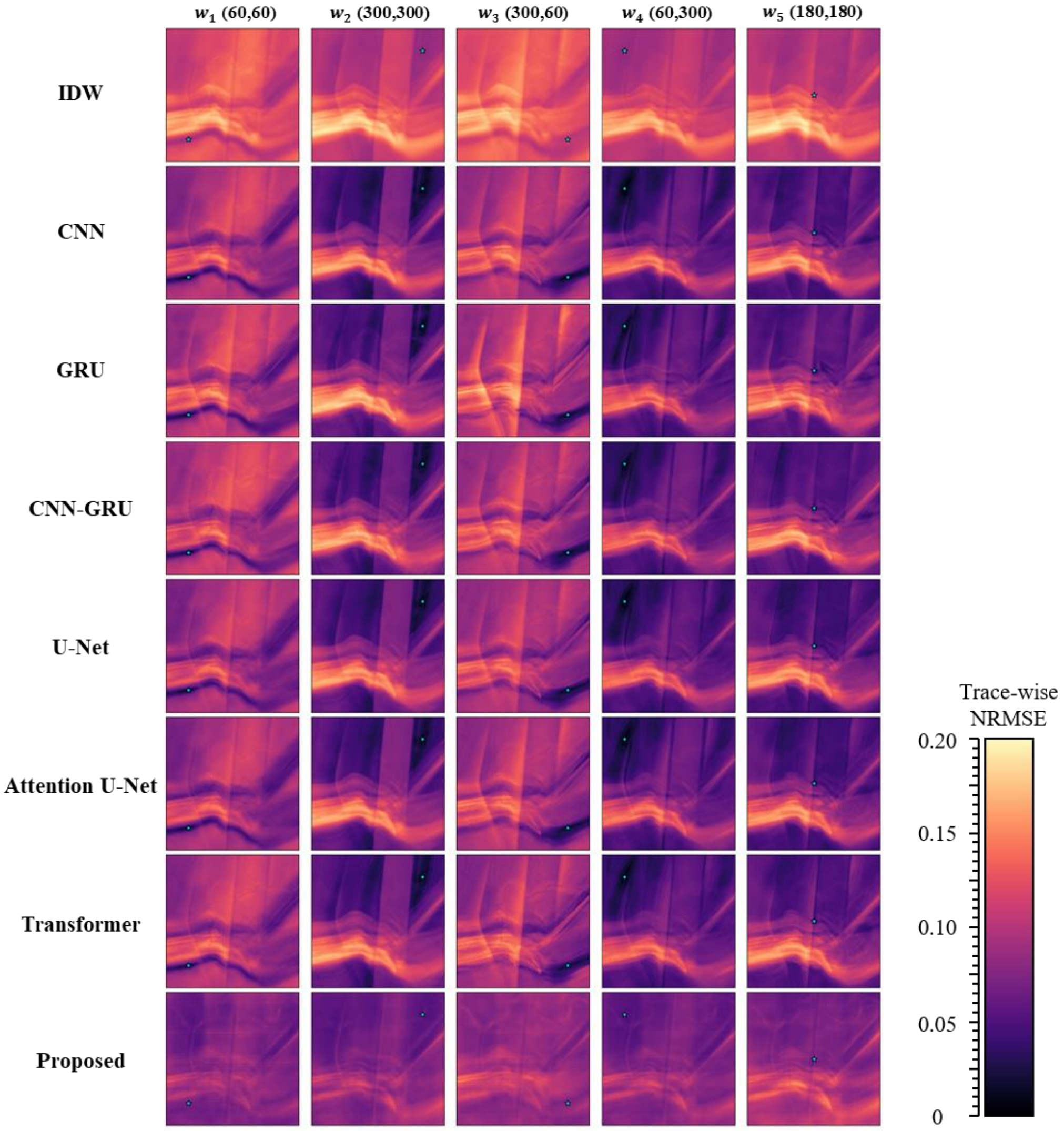

Fig. 5. Trace-wise NRMSE maps for the Overthrust experiments. Columns represent the five single-well configurations, and rows represent the IDW prior, six target-trained baselines, and the proposed method. Stars indicate the conditioning-well locations.

Table 1. Full-volume Overthrust results across the five single-well configurations. All metrics are reported as mean ± standard deviation.

| Method | NRMSE ↓ | P95 ↓ | PCC ↑ |
|---|---|---|---|
| IDW prior | 0.1209 ± 0.0065 | 0.1632 ± 0.0062 | 0.9039 ± 0.0116 |
| CNN | 0.0844 ± 0.0111 | 0.1337 ± 0.0086 | 0.9559 ± 0.0123 |
| GRU | 0.0878 ± 0.0132 | 0.1393 ± 0.0109 | 0.9525 ± 0.0132 |
| CNN-GRU | 0.0858 ± 0.0116 | 0.1342 ± 0.0093 | 0.9542 ± 0.0128 |
| U-Net | 0.0831 ± 0.0098 | 0.1310 ± 0.0102 | 0.9571 ± 0.0106 |
| Attention U-Net | 0.0834 ± 0.0095 | 0.1320 ± 0.0098 | 0.9569 ± 0.0102 |
| Transformer | 0.0838 ± 0.0099 | 0.1331 ± 0.0083 | 0.9564 ± 0.0111 |
| Proposed | **0.0814 ± 0.0057** | **0.1074 ± 0.0093** | **0.9602 ± 0.0036** |

### 3.4 Field-data application

The processed field seismic volume contains $250 \times 500 \times 400$ samples with a temporal sampling interval of 1 ms, corresponding to 125000 seismic traces. The inversion evaluation focuses on the 0.85-1.15 s interval. Four wells are available within the survey. One well was designated as the conditioning well, and its impedance log was low-pass filtered and interpolated along interpreted horizons to construct the low-frequency prior. The remaining three wells were excluded from prior construction, output normalization, and model selection and were used exclusively for blind evaluation.

This setting represents an extremely sparse deployment scenario in which only one target-area well is available. The U-Net was trained exclusively using the conditioning trace, with its architecture and hyperparameters fixed from the preceding Overthrust experiments. No additional well was used for validation because its impedance labels would influence model selection, provide additional target-area information, and prevent that well from being treated as blind.

The seismic volume and low-frequency prior were processed using the normalization and overlapping-block reconstruction procedures described in Sections 2.7 and 3.1. All offline-trained components remained fixed during field application. Once the frozen prediction for the complete survey had been obtained, the mean and standard deviation of the prediction and measured impedance at the conditioning well were used to determine the two global affine normalization parameters defined in Section 2.7.3. The same parameters were then applied uniformly to the complete predicted volume without neural-network optimization.

Figure 6 presents the seismic data, one-well low-frequency prior, and inversion results. Figure 7 compares the impedance curves at the three blind wells, and the corresponding metrics are reported in Table 2.

Figure 6 shows that the proposed method recovers laterally varying impedance features that are absent from the one-well low-frequency prior. The blind-well comparisons in Fig. 7 further show improved agreement with the measured impedance curves at all three locations.

As shown in Table 2, the proposed method achieved the lowest NRMSE and highest PCC at each of the three blind wells. Its mean NRMSE was 0.1260, representing reductions of 20.30% and 20.95% relative to the low-frequency prior and U-Net, respectively. Its mean PCC reached 0.8982, compared with 0.8216 for the prior and 0.7421 for U-Net.

Because the global affine normalization has a positive scale factor, it does not alter PCC. The consistently higher correlations therefore originate from the spatial and temporal impedance information contained in the frozen prediction, while the conditioning-well normalization aligns its global impedance range. These results demonstrate that useful inversion information can be transferred beyond the conditioning location without target-specific neural-network optimization.

Table 2. Quantitative comparison of field inversion results at three blind wells.

| | Well 1 | | Well 2 | | Well 3 | | Mean | |
|---|---|---|---|---|---|---|---|---|
| | NRMSE | PCC | NRMSE | PCC | NRMSE | PCC | NRMSE | PCC |
| Low frequency prior | 0.1844 | 0.7846 | 0.1560 | 0.8691 | 0.1339 | 0.8110 | 0.1581 | 0.8216 |
| U-Net | 0.2092 | 0.6188 | 0.1421 | 0.8212 | 0.1268 | 0.7862 | 0.1594 | 0.7421 |
| Proposed | **0.1403** | **0.8555** | **0.1321** | **0.9190** | **0.1055** | **0.9201** | **0.1260** | **0.8982** |

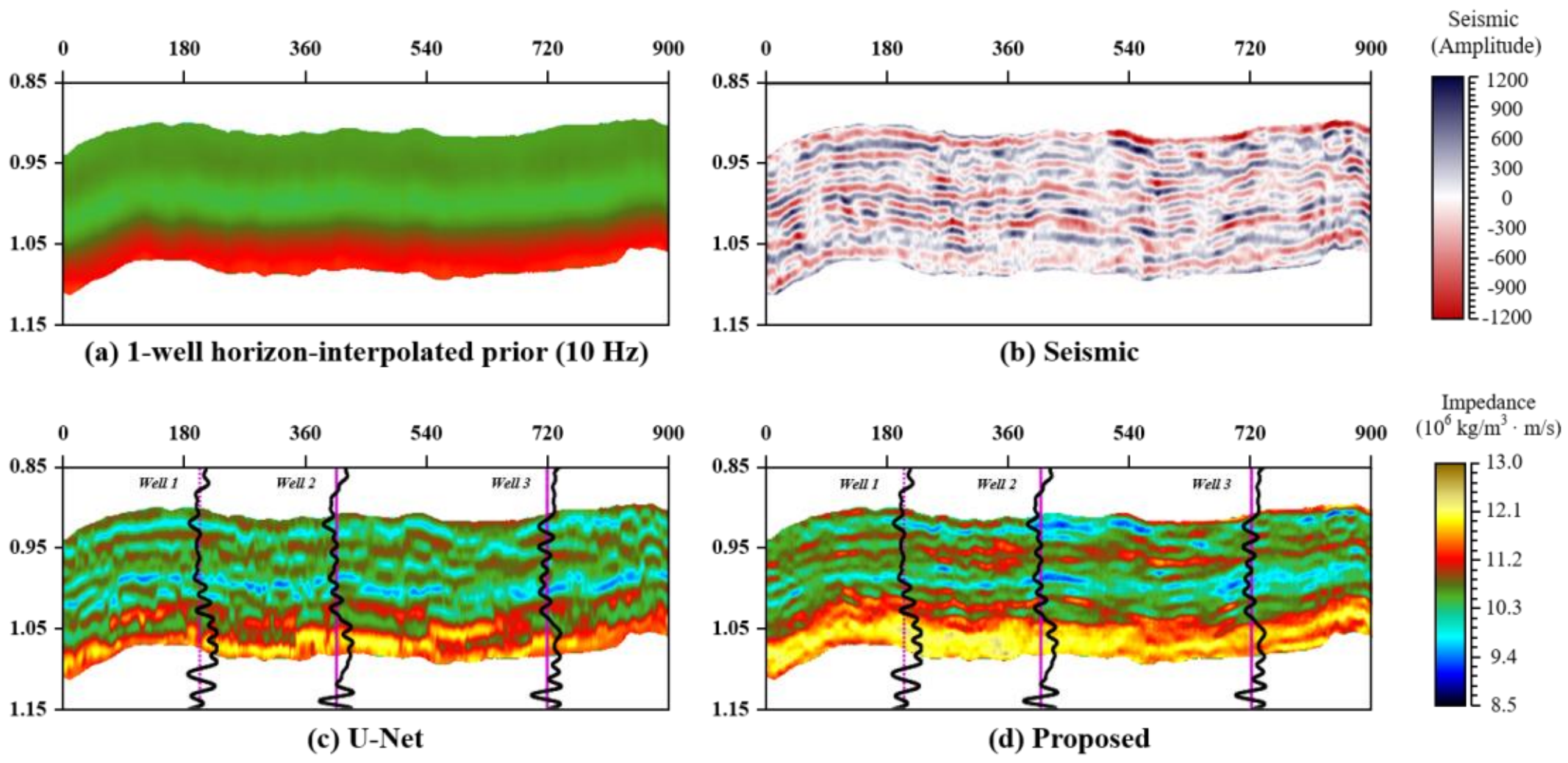

Fig. 6. Field inversion results using a one-well horizon-interpolated low-frequency prior. (a) One-well 10-Hz low-frequency prior; (b) seismic data; (c) U-Net; and (d) proposed method.

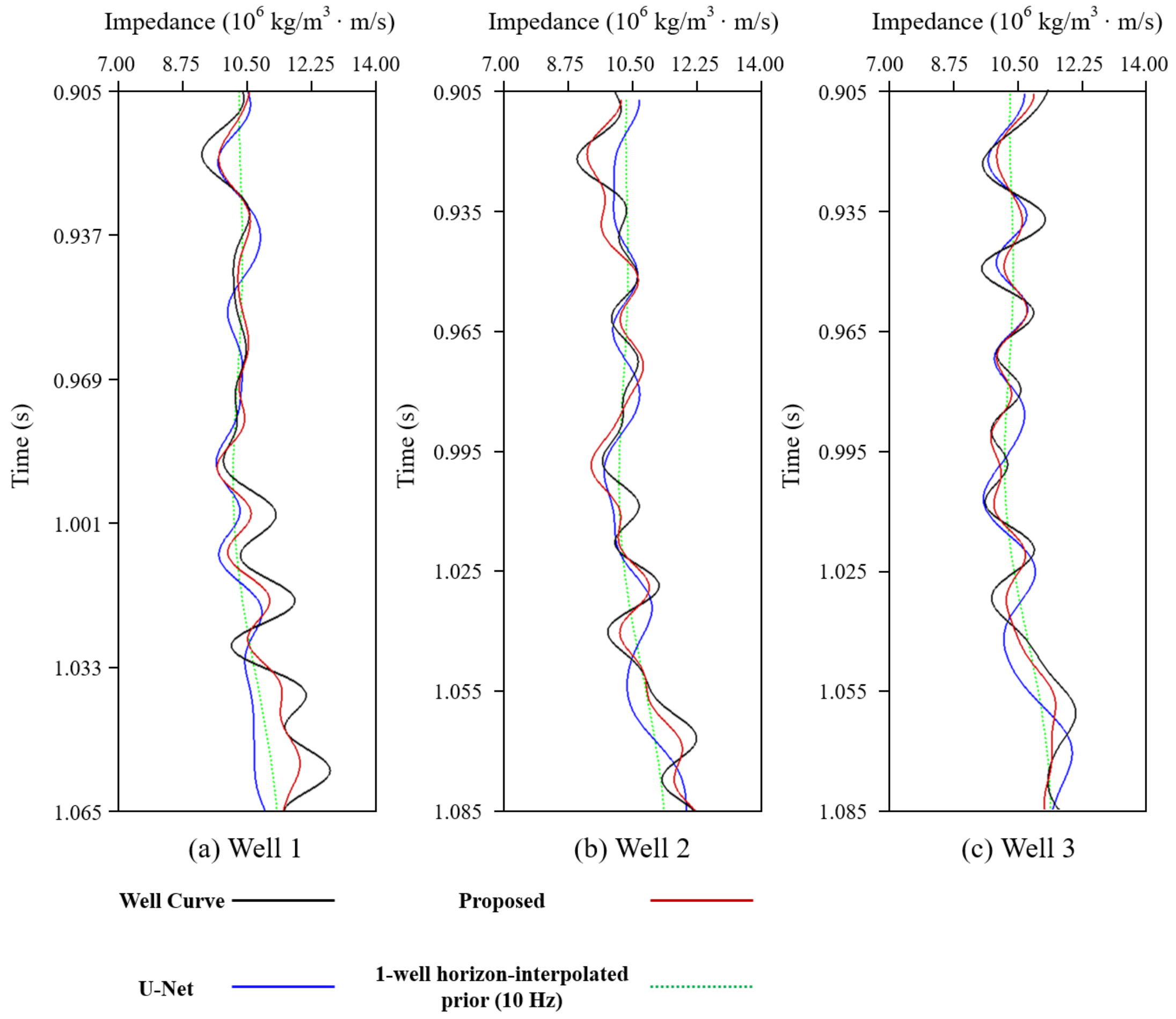

Fig. 7. Blind-well impedance comparisons. (a) Well 1; (b) Well 2; and (c) Well 3.

## 4. Discussion

### 4.1 Influence of low-frequency prior representativeness

The effectiveness of sparse-well inversion depends not only on the number of conditioning wells but also on whether their impedance trends are representative of the target area. In the single-well Overthrust experiments, inverse-distance weighting replicates one filtered log across the lateral plane and therefore introduces no lateral structural variation. Nevertheless, these priors retain a relatively high mean PCC of 0.9039 with the target volume because the Overthrust model contains a strong and laterally consistent vertical impedance trend. Under this condition, the target-trained baselines receive an informative background model and only need to learn a comparatively small residual correction. This partly explains why the proposed method reduces the mean NRMSE by only approximately 2.04% relative to U-Net.

The benefit of the transferred representation is particularly evident in the $W_3$ configuration, where the IDW prior has the lowest PCC (0.8859) among the five layouts. U-Net obtains an NRMSE of 0.0958, a P95 trace-wise NRMSE of 0.1256, and a PCC of 0.9434 whereas the proposed method achieves 0.0883, 0.1120, and 0.9536, respectively. This corresponds to reductions of 7.82% in NRMSE and 10.83% in P95 relative to U-Net. The substantially larger improvement in this configuration shows that the transferred three-dimensional representation becomes particularly valuable when the conditioning log is less representative of the surrounding volume.

The field results reinforce this interpretation. U-Net, trained exclusively on the single conditioning trace, performs worse on average than the unchanged low-frequency prior, with an NRMSE of 0.1594 versus 0.1581 and a PCC of 0.7421 versus 0.8216. This behavior is consistent with the learned residual correction overfitting the local seismic–impedance relationship and introducing inaccurate modifications where that relationship changes laterally. In comparison, the proposed method maintains a mean NRMSE of 0.1260 and a PCC of 0.8982 across the blind wells. Because affine normalization does not alter PCC, the higher correlation reflects spatial and temporal information contained in the frozen prediction. Together with the absence of target-specific network retraining, these results demonstrate its practical value under extremely sparse well control.

### 4.2 Transferable prediction and conditioning-well normalization

Cross-survey differences in impedance range can introduce a systematic amplitude bias even when the main impedance variations are preserved. Conditioning-well normalization addresses this global scale-and-offset mismatch without updating the network or learning a spatially varying seismic-impedance relationship.

A positive affine transformation can reduce amplitude error but cannot change PCC, recover missing lateral structures, or alter the alignment of impedance variations. The PCC values of 0.9602 for Overthrust and 0.8982 for the field survey therefore reflect information contained in the frozen prediction rather than the normalization procedure. The normalization only expresses this prediction in an impedance range consistent with the target survey.

The same procedure was applied to all learned methods using the corresponding conditioning log. Consequently, the reported differences reflect the quality of the underlying predictions rather than unequal amplitude treatment. This provides a transparent and inexpensive way to deploy the transferred network under extremely sparse well control.

### 4.3 Roles of the main model components

Table 3 summarizes the ablation results under the five single-well configurations. All predictions were generated using the same overlap and conditioning-well normalization settings. The complete model achieves the best mean performance across all three metrics.

Removing Stage-I seismic pretraining increases NRMSE from 0.0814 to 0.0857 and P95 from 0.1074 to 0.1177. This indicates that masked seismic pretraining improves the transferability of the three-dimensional seismic representation and reduces both average and localized errors.

Removing gated prior conditioning results in the lowest PCC and the highest P95

among the evaluated variants. Although the low-frequency prior remains the explicit output background, it is no longer injected into the decoder as conditional features. The resulting degradation indicates that input-dependent fusion helps the network use sparse low-frequency information more effectively when its lateral reliability varies.

Frequency consistency has a smaller influence on mean NRMSE and PCC, but its removal increases P95 from 0.1074 to 0.1176. Its main contribution is therefore to suppress localized large errors rather than determine the overall inversion accuracy. Taken together, the ablations identify seismic pretraining and gated prior conditioning as the primary components supporting cross-survey transfer, while frequency consistency provides an additional improvement in spatial stability.

Table 3 Ablation results averaged over the five single-well configurations. The values after ± denote standard deviations across well locations.

| Model configuration | NRMSE | P95 | PCC |
|---|---|---|---|
| Full model | **0.0814 ± 0.0057** | **0.1074 ± 0.0093** | **0.9602 ± 0.0036** |
| w/o Stage-I seismic pretraining | 0.0857 ± 0.0057 | 0.1177 ± 0.0100 | 0.9580 ± 0.0041 |
| w/o gated prior conditioning | 0.0841 ± 0.0069 | 0.1186 ± 0.0076 | 0.9573 ± 0.0080 |
| w/o frequency consistency | 0.0821 ± 0.0056 | 0.1176 ± 0.0111 | 0.9601 ± 0.0034 |

### 4.4 Practical scope and limitations

The proposed framework is intended for post-stack impedance inversion when target-survey well information is extremely limited and retraining a three-dimensional network is impractical. It is not intended to replace survey-specific supervised inversion when numerous representative wells and sufficient computational resources are available. Its main value lies in transferring offline-learned seismic and inversion information to a new survey while restricting target-dependent processing to low-frequency prior construction and conditioning-well normalization.

As with other pretrained models, the frozen prediction depends on the geological and seismic variability represented during offline training. A sufficiently large distribution difference may introduce errors that cannot be removed completely by the available target information. The normalization used here corrects only a global scale and offset. This limited form is intentional: it is transparent, requires no gradient-based optimization, and cannot generate lateral structures from the conditioning well. Consequently, the improved blind-well correlation must originate from the frozen three-dimensional prediction rather than from the normalization procedure. More complex spatially varying discrepancies remain outside the role of this adjustment.

Stage-II seismic synthesis is based on the post-stack convolutional model, and the auxiliary branch estimates an equivalent Ricker dominant frequency. The estimate serves as a consistency diagnostic rather than a complete field-wavelet solution. More complex nonstationary wavelet parameterizations may be considered in future extensions.

The low-frequency prior also remains dependent on the available well log and the selected interpolation procedure. Training with imperfect sparse-well priors and using gated conditioning reduce sensitivity to prior uncertainty but do not assume that every prior is equally informative. The field experiment further supports the practical value of the framework under one-well control. The frozen network preserves useful lateral impedance information beyond the conditioning location, while simple normalization reduces systematic amplitude differences at a low deployment cost. This provides an efficient route for cross-survey inversion without retraining the three-dimensional model.

## 5. Conclusion

This study developed a two-stage offline pretraining framework for cross-survey impedance inversion under extremely sparse well control. Stage I learns three-dimensional seismic representations through masked reconstruction, and Stage II learns sparse-prior-conditioned inversion from synthetic geological models. During deployment, the three-dimensional network remains frozen; target wells are used only to construct the low-frequency prior and determine two normalization parameters.

Across the five single-well Overthrust configurations, the proposed method achieved a mean NRMSE of 0.0814, lower than 0.0831 for the best target-trained baseline, and substantially reduced the P95 trace-wise NRMSE by 18.02%, from 0.1310 to 0.1074. It also obtained the highest mean PCC of 0.9602. These results show that the frozen network achieves competitive average accuracy while producing fewer localized large errors without target-specific network training. Ablation experiments identified seismic pretraining and gated prior conditioning as the principal contributors, whereas dominant-frequency consistency mainly reduced the upper tail of the spatial error distribution.

In the field experiment with one conditioning well, the proposed method achieved a mean NRMSE of 0.1260 and a mean PCC of 0.8982 at three blind wells, outperforming the U-Net at every blind location. Because positive affine normalization does not alter PCC, the consistently high correlation reflects lateral impedance information contained in the frozen prediction. The framework therefore provides a practical, low-cost solution for cross-survey inversion under extremely sparse well control without retraining the three-dimensional network.